\documentclass[lettersize,journal]{IEEEtran}
\usepackage{amsmath,amsfonts}
\usepackage{algorithmic}
\usepackage{algorithm}
\usepackage{array}
\usepackage[caption=false,font=normalsize,labelfont=sf,textfont=sf]{subfig}
\usepackage{textcomp}
\usepackage{stfloats}
\usepackage{pifont}
\usepackage{url}
\makeatletter

\newcommand{\Rmnum}[1]{\expandafter\@slowromancap\romannumeral #1@}
\makeatother
\usepackage{graphicx}
\usepackage{multirow}
\usepackage[table,xcdraw]{xcolor}
\usepackage{verbatim}
\usepackage{graphicx}
\usepackage{cite}
\usepackage{booktabs}
\usepackage[section]{placeins}

\begin{document}

\title{Human-centric Behavior Description in Videos: \\New Benchmark and Model}

\author{Lingru Zhou,
        Yiqi Gao,
        Manqing Zhang,
        Peng Wu*,
        Peng Wang*,
        and Yanning Zhang
\thanks{ Lingru Zhou, Yiqi Gao, Peng Wu, Peng Wang and Yanning Zhang are with the School of Computer Science and Ningbo Institute, Northwestern Polytechnical University, Xi’an, China. Manqing Zhang is with the School of Software, Northwestern Polytechnical University, Xi’an, China.}
\thanks{* The corresponding authors are Peng Wu (xdwupeng@gmail.com) and Peng Wang (peng.wang@nwpu.edu.cn).}}

\markboth{Journal of \LaTeX\ Class Files,~Vol.~14, No.~8, August~2023}%
{Shell \MakeLowercase{\textit{et al.}}: A Sample Article Using IEEEtran.cls for IEEE Journals}



\maketitle

\begin{abstract}
In the domain of video surveillance, describing the behavior of each individual within the video is becoming increasingly essential, especially in complex scenarios with multiple individuals present. This is because describing each individual's behavior provides more detailed situational analysis, enabling accurate assessment and response to potential risks, ensuring the safety and harmony of public places. Currently, video-level captioning datasets cannot provide fine-grained descriptions for each individual's specific behavior. However, mere descriptions at the video-level fail to provide an in-depth interpretation of individual behaviors, making it challenging to accurately determine the specific identity of each individual. To address this challenge, we construct a human-centric video surveillance captioning dataset, which provides detailed descriptions of the dynamic behaviors of 7,820 individuals. Specifically, we have labeled several aspects of each person, such as location, clothing, and interactions with other elements in the scene, and these people are distributed across 1,012 videos. Based on this dataset, we can link individuals to their respective behaviors, allowing for further analysis of each person's behavior in surveillance videos. Besides the dataset, we propose a novel video captioning approach that can describe individual behavior in detail on a person-level basis, achieving state-of-the-art results. To facilitate further research in this field, we intend to release our dataset and code.
\end{abstract}


\section{Introduction}
\begin{figure}[!t]
\centering
\includegraphics[width=3.5in]{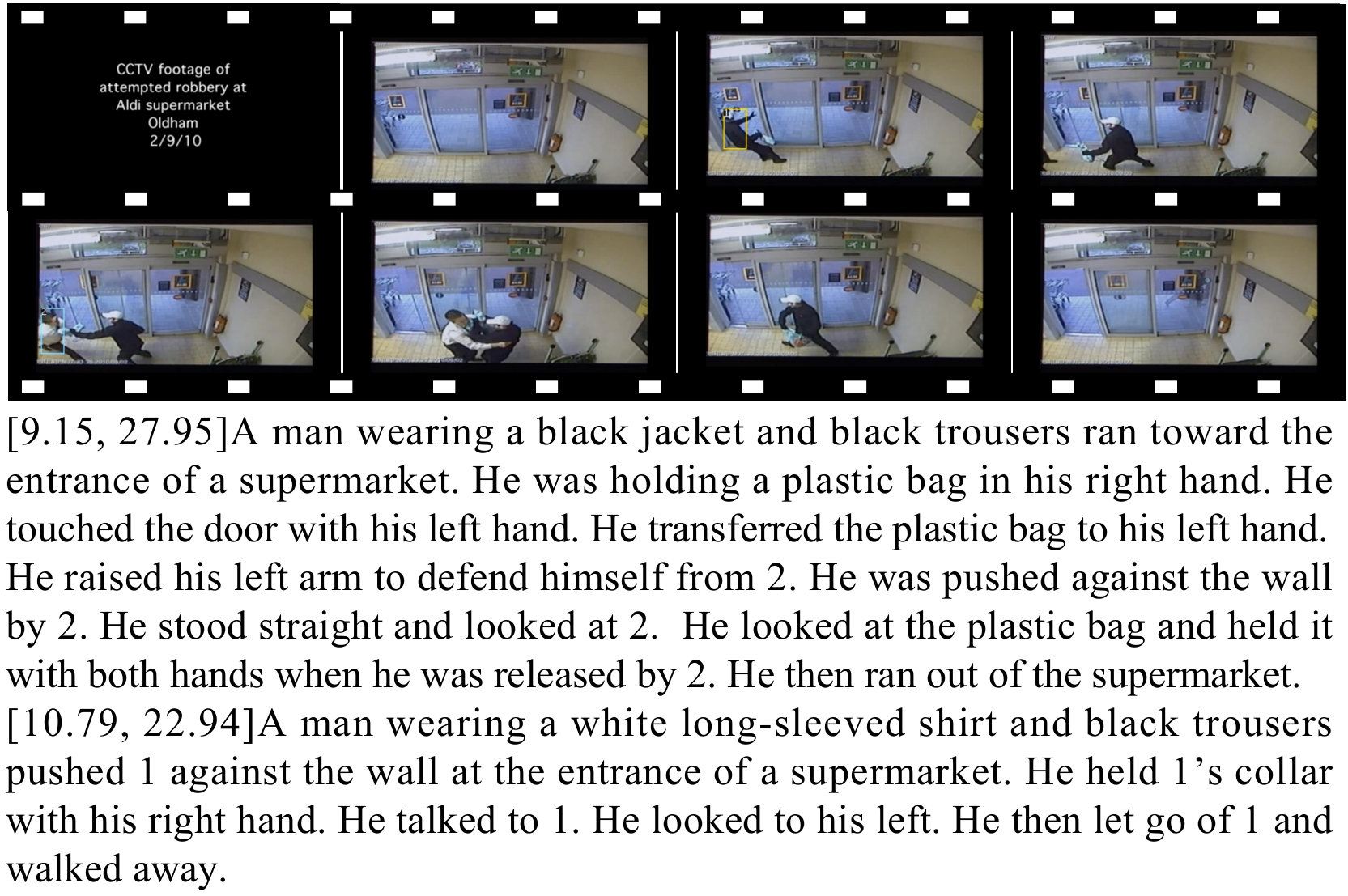}
\caption{An example from our collected UCF-crime captioning dataset. To aid in-depth research, we have annotated the bounding box for the first frame in which each individual appears and provided an objective description of the behavior of each person appearing in our dataset.}
\label{fig:example}
\end{figure}

\begin{figure*}[!t]
\centering
\includegraphics[width=7in]{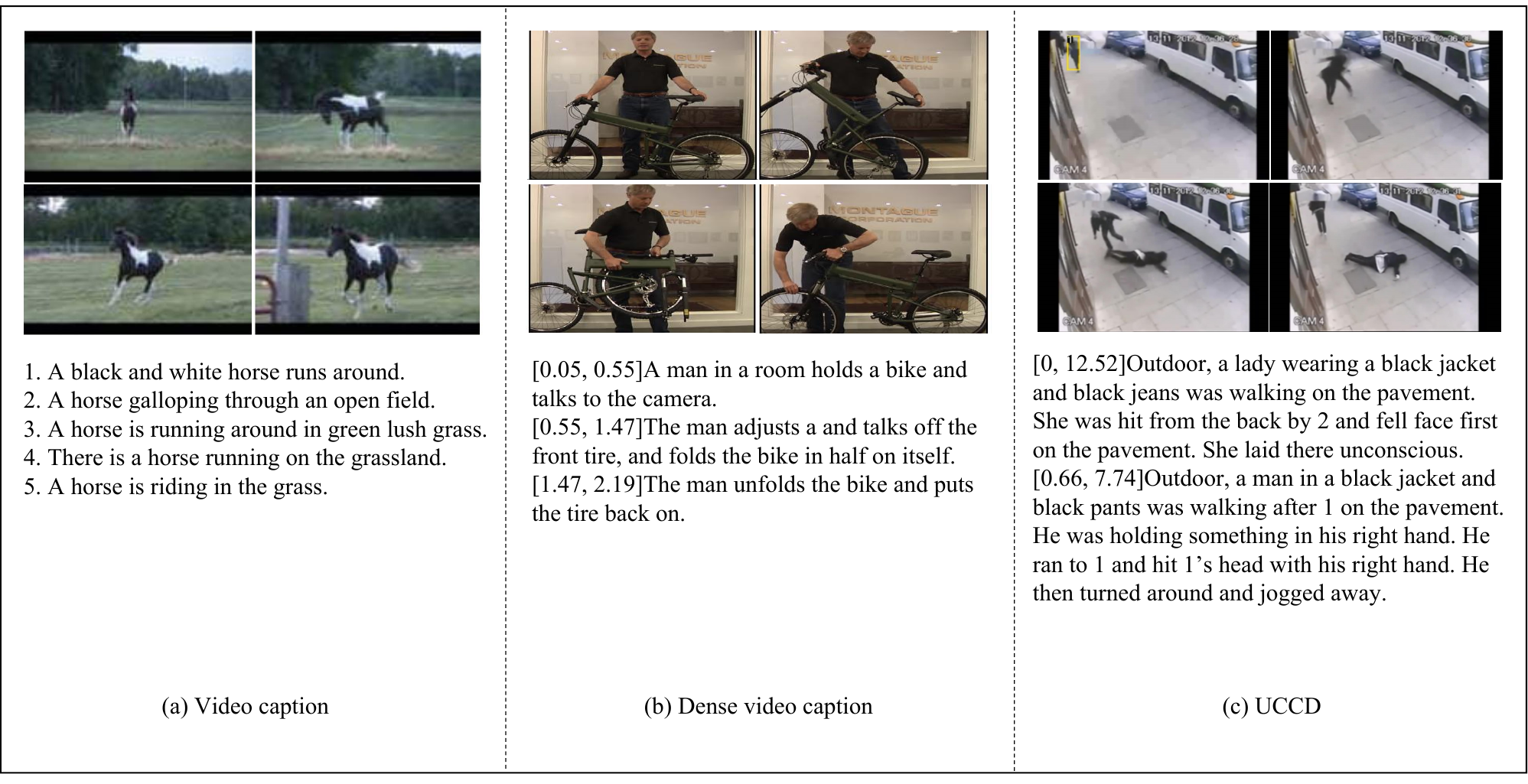}
\caption{The figure are respectively examples of the video caption, dense video caption, and UCF-crime caption. The video caption, which was proposed first, is a description of the video. The dense video caption divides a video into several time frames, assigning temporal segmentation to events, and then describing each event. The UCF-crime caption identifies the time period when a person appears and disappears, and describes the behavior of each person within this period.}
\label{fig:Intro}
\end{figure*}

\IEEEPARstart{W}{ith} the rapid development of security technology, vision applications based on video surveillances have become the focus of many scholars and the industrial community~\cite{ref45}. So far, research and datasets related to surveillance videos mainly focus on anomaly detection~\cite{ref38,ref39,ref40,ref41,ref62,ref63,ref64,ref65}. Undoubtedly, this is an important research field; however, it overlooks some more complex scenarios. For example, in the real world, we need not only to detect abnormal events but also to analyze individual abnormal behaviors in surveillance videos~\cite{ref44,ref52} to prevent the occurrence of abnormal events or stop ongoing criminal activities from worsening. Existing research does not fully encompass these scenarios because they demand a human-centric behavioral video captioning dataset to describe individuals, aiding the analysis of individual behaviors, a simple example of which is illustrated in Figure~\ref{fig:example}. Currently available datasets, as shown in Figure~\ref{fig:Intro}, mainly describe entire videos or divide videos into several events for description, failing to meet this demand. Obtaining this data requires a large amount of human resources, posing significant challenges in data collection.

\begin{figure}[!t]
\centering
\includegraphics[width=3.5in]{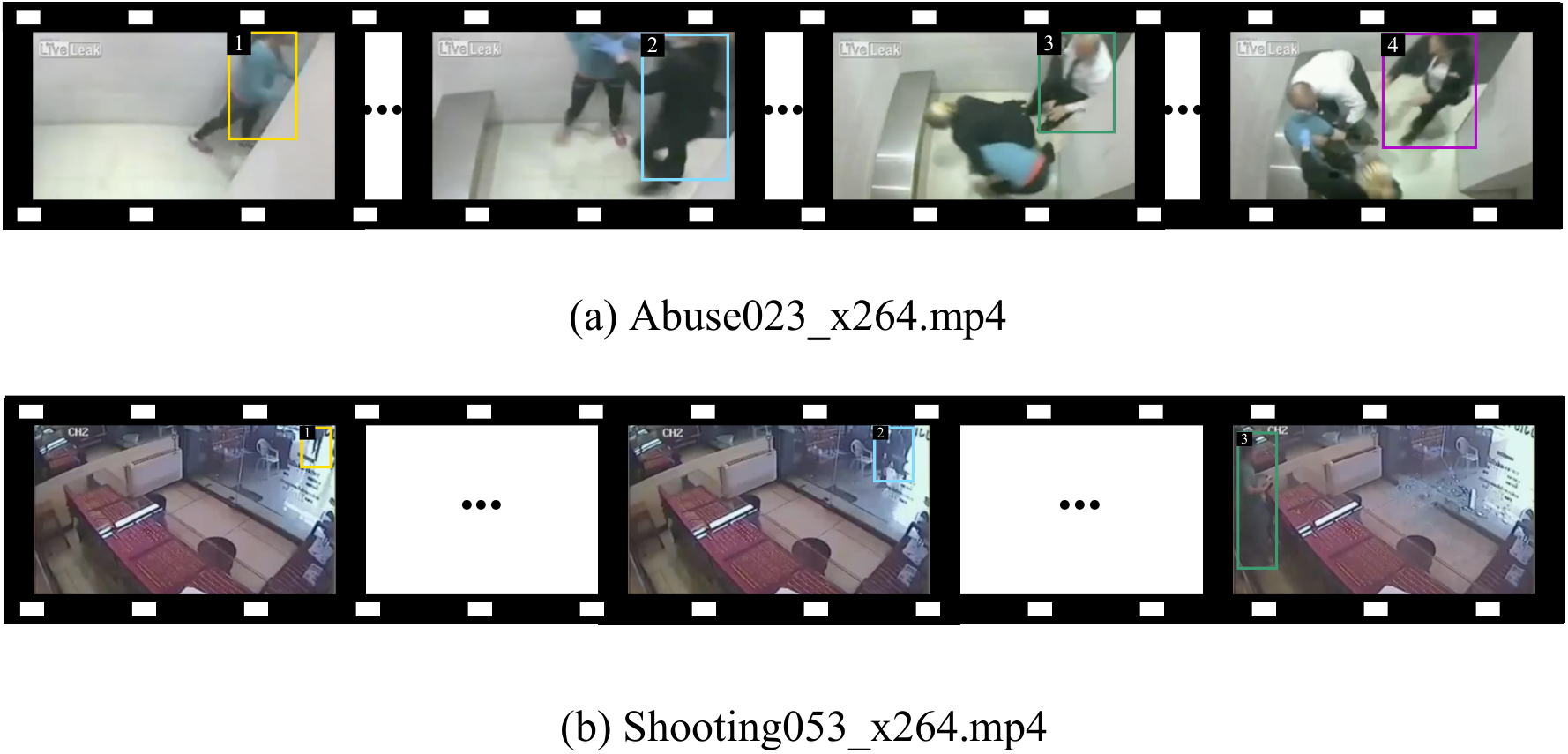}
\caption{Subfigure (a) and (b) are videos under different scenarios. In both videos, different individuals are marked in sequence of appearance with vivid yellow, sky blue, emerald green, purple, etc.}
\label{fig:bboxcolor}
\end{figure}

To address this issue, we propose a human-centric behavioral video captioning dataset: the UCF-crime captioning dataset (UCCD) as shown in Figure~\ref{fig:example}. The UCCD dataset includes 1,012 videos and descriptions of the behavior of 7820 individuals, covering various scenarios, including normal and abnormal events. It not only solves the problem of analyzing individual behaviors in surveillance videos but also contains some unique features. For each individual in a video, we detect them in the frame where they first appear and mark them with different colored bounding boxes, tracking them until they disappear. Figure~\ref{fig:bboxcolor} represents an example of the color of the boundary boxes appearing in the same order in different videos. This not only enhances the captioning task but also provides a more diversified dataset for research. 

Based on the rich annotation information of our dataset, we propose a human-centric video captioning method. This method extracts frame and human features from the deformable transformer and generates a caption for each individual's behavior in the video through the localization head and captioning head. By introducing a module for person detection and tracking, our method can accurately divide the video into different segments according to different people, thereby fully understanding the video content while avoiding information omission and repeated caption generation caused by unreliable estimates of the number of people.

The contributions of this paper are three-fold:

\textbullet~We construct a human-centric video surveillance captioning dataset across 1,012 videos, detailing the dynamic behaviors of 7,820 individuals. Specifically, we label various aspects such as location, clothing, and interactions with other elements in the scene, greatly enriching the comprehension of human interactions in complex scenarios.

\textbullet~To the best of our knowledge, we first propose the video surveillance captioning task, enabling the understanding of human actions in videos and the output of descriptions of human behavior, marking a novel direction in the field of video surveillance.

\textbullet~We propose a novel video captioning approach that is designed on a person-level basis, achieving state-of-the-art results.




The rest of our paper is organized as follows: In section \Rmnum{2}, we introduce the relevant work concerning video captioning datasets, video captioning methods, and video anomaly detection. In section \Rmnum{3}, we introduce the UCCD dataset. In section \Rmnum{4}, we propose a new method for captioning individual behavior in video surveillance s. In section \Rmnum{5}, we validate that our proposed method is superior to existing methods, emphasizing the value of the UCCD dataset in the development of advanced video captioning and surveillance technologies. In section \Rmnum{6}, we conclude the paper.


\section{RELATED WORK}
{\bf{Video Captioning Datasets.}} In recent years, a number of datasets have been compiled for video captioning research\cite{ref1,ref2,ref3,ref4,ref5,ref50}. These datasets vary in size, scope, and focus, but they all provide valuable data for training and evaluating video captioning models. With MSR-VTT~\cite{ref1}, researchers were offered a large-scale video dataset named Microsoft Research Video Description Dataset. It includes around 10,000 video clips and 200,000 text descriptions spanning approximately 200 categories, gathered from online video sharing platforms. Similarly, the HowTo100M dataset~\cite{ref2} provides an extensive dataset of around 1.36 million YouTube videos with their corresponding spoken descriptions, serving approximately 1.3 billion video clips. The main aim of this dataset is to facilitate "how-to" tasks such as action recognition, object recognition, and scene recognition.

Likewise, ActivityNet Captions~\cite{ref3} was constructed as a dataset encompassing about 20,000 YouTube videos and their language descriptions, targeting the temporal context~\cite{ref50} description of videos. The YouCook2 dataset~\cite{ref4}, focusing on cooking videos, comprises approximately 2000 YouTube cooking videos with paragraph-level descriptions. Microsoft Video Description Corpus (MSVD)~\cite{ref5}, offers around 2000 YouTube short videos with descriptions in multiple languages. In contrast, the WebVid~\cite{ref6} dataset consists of about half a million videos scraped from the internet along with their annotations.

Providing bilingual descriptions, the VATEX~\cite{ref7} incorporates around 41,250 videos. The TGIF dataset~\cite{ref8} encompasses around 100,000 GIF animations with their English descriptions. A unique benchmarking dataset, VALUE~\cite{ref9}, is designed for evaluating multimodal (visual + language) AI systems.

In the realm of video question answering tasks, HowTo2QA~\cite{ref10} and MTLAQA datasets~\cite{ref11} have been developed. The former contains "How-to" type videos with corresponding questions and answers while the latter is a multi-task learning question answering dataset, covering various task types such as action recognition, object detection, and more.

Lastly, the TVC (TV show Clip Captioning Dataset)~\cite{ref12} is a video captioning dataset encompassing around 15,000 TV show clips with their descriptions. Advancing beyond vision, the VALOR-1M dataset was developed by Chen et al.~\cite{ref28}. This dataset comprises of 1 million video clips derived from AudioSet~\cite{ref29}, each paired with annotated audiovisual captions.

{\bf{Video Captioning Methods.}} In the realm of urban surveillance, significant advancements have been made in action recognition, object tracking, and video captioning. Krishna et al.~\cite{ref13} pioneered the multifaceted task of video captioning with a dense model, integrating a multi-scale proposal~\cite{ref60} module for localization and an attention-driven LSTM for context-aware caption generation~\cite{ref14}\cite{ref15}. This innovation has sparked further developments, such as the research by Ghaderi et al.~\cite{ref18} introduced a temporal-spatial attention module to improve the accuracy of action recognition. In contrast, Wang et al.~\cite{ref19} presented a comprehensive multi-stage framework that prioritizes precise action identification. The exploration of synergies between video captioning sub-tasks has also provided valuable insights. For example, Li et al.~\cite{ref16} introduced a proxy task to predict language rewards of generated sentences, optimizing the localization module. In a similar vein, Wang et al.~\cite{ref17} presented PDVC, a model that capitalizes on inter-task interactions by sharing intermediate features.
Building on these advances, we propose a novel task: generating sequential captions for each individual throughout a video, thereby unifying the fields of action recognition, object tracking, and video captioning into a holistic, individual-centric approach.

{\bf{Video Anomaly Detection.}} Video Anomaly Detection (VAD) refers to the identification and detection of events deviating from normal behaviors, widely applied in video surveillance  scenarios. Depending on the developmental stages of the algorithms, they can be classified into three categories: traditional machine learning methods, hybrid methods of traditional machine learning and deep learning, and deep learning methods. Most studies employ traditional handcrafted features, such as histogram of oriented gradients (HOG)~\cite{ref69}, histogram of optical flow (HOF)~\cite{ref85}, local binary pattern (LBP)~\cite{ref70}, etc., to represent crowd appearance and motion information, then detect anomalies using conventional machine learning techniques. Given that deep features exhibit stronger descriptive ability than handcrafted features, during the hybrid phase, algorithms use deep features to replace handcrafted ones, followed by anomaly detection using traditional machine learning methods. Discrimination models that saw significant advancements during this stage mainly include point models, and applications focused primarily on cluster discrimination~\cite{ref71,ref72,ref73}, reconstruction discrimination~\cite{ref74,ref75}, and others~\cite{ref76}. In the phase of deep learning methods, algorithms combine feature extraction steps with model training steps, conducting anomaly detection through end-to-end methods~\cite{ref62,ref77,ref78,ref79,ref80,ref81,ref82}. Currently, video anomaly detection faces multiple challenges, such as the vagueness in anomaly event definition, the lack of clear delineation between normal and abnormal samples~\cite{ref67}, and the scene-dependence of anomaly event definition~\cite{ref68}. The same event under different scenes may present different anomalous attributes. In response to these issues, we propose a new task, which includes captioning videos under normal and abnormal scenarios, offering a new foundation and direction for video anomaly detection.

\section{DATASET}
In this section, we provide the UCF-crime captioning dataset, focusing on the data collection process and annotations. The UCF-crime dataset serves as the foundation for our work, where we have augmented it with detailed captions for each individual. 

\begin{figure}[htbp]
\centering
\includegraphics[width=3.5in]{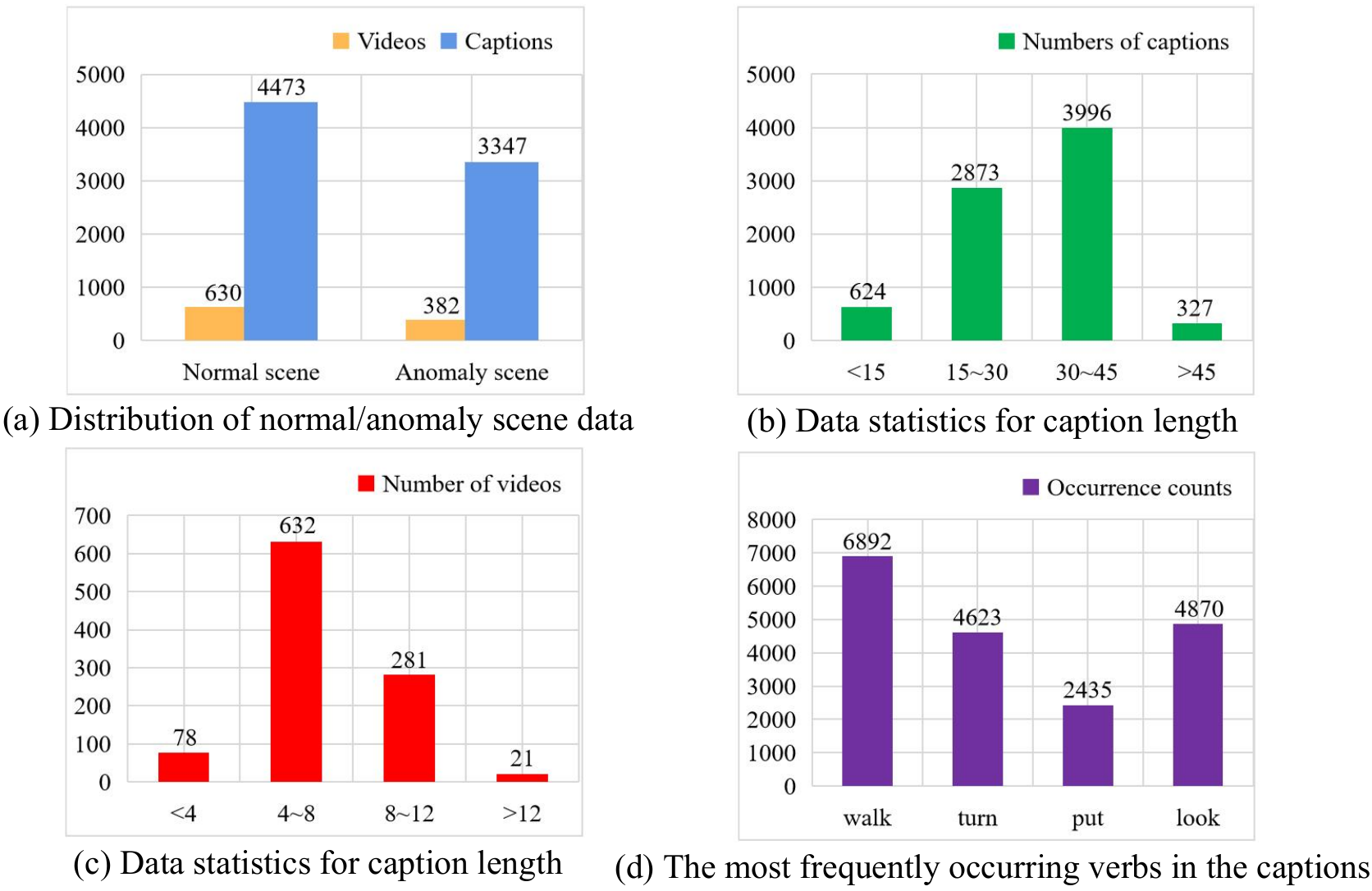}
\caption{Statistical information on normal and anomaly scene data, caption length, number of people appearing in each video, and the four verbs that appear most frequently in captions.}
\label{fig:dataset}
\end{figure}

As shown in Figure \ref{fig:dataset}, we conducted a detailed analysis of the UCCD dataset. In the dataset, the number of videos in normal scenes far exceeds that of abnormal scenes. The length of the captions mainly ranges from 30 to 45 words, with an average of 34 words per caption. The number of people appearing in the videos is usually between 4 to 8, but in some abnormal scenes, such as fights, road accidents, and attacks, the number of people can increase, with some videos even featuring more than 20 people. As the majority of videos are under normal scenes, some commonly performed physical actions appear frequently in the captioning of people's behavior, such as "walk", "turn", and "look".

\subsection{Data Collection}
In our endeavor, we engaged 20 native speakers to manually caption the videos, dedicating a substantial 200 hours to provide targeted training prior to formal annotation. During the labeling process, each video required simultaneous annotation by five individuals. The specific tasks involved first detecting individuals present in the UCF-crime dataset, then marking a bounding box on the first frame of their appearance in the video, and assigning them consecutive numbers based on the order of their appearances. Following this, we provided an objective description of their actions throughout the process from their appearance to their disappearance in the video. To ensure the accuracy of the caption and the objectivity of the action description, once the five individuals had annotated each video, a seasoned caption annotation expert would sift through their work, selecting the caption most fitting to the setting. After over 5,000 hours of annotation, we ultimately collected 1,012 videos with a total duration of 112 hours, containing a total of 7,820 captions, at a cost of approximately 6,017 dollars.
\begin{table*}[htbp]
\centering
\caption{Comparison with other datasets. we compare UCCD with existing video caption datasets in several aspects: domain, video source, average time, caption length, caption target, and target type.We discovered that the UCCD dataset is a video caption dataset based on human-centric units, originating from video surveillance s.}
\label{tab:dataset}
\setlength{\tabcolsep}{3.15mm}
{%
\begin{tabular}{l|c|c|c|c|c|c}
\hline
Datasets             & Domain & Video Source       & Average Time & Caption Length & Caption Target & Target Type            \\ \hline
MSR-VTT~\cite{ref1}             & Open    & YouTube            & 14.8s        & 9              & video          & generic event          \\
MSVD~\cite{ref5}                 & Open    & YouTube            & 9.0s         & 8              & segment        & generic event          \\
VATEX~\cite{ref7}                & Open   & YouTube            & 10.0s        & 15             & video          & action                 \\
ActivityNet Captions~\cite{ref3} & Action   & YouTube            & 120.0s       & 13.5           & segment        & action                 \\
\hline
UCCD                 & Security    & Video surveillance & 42.3s        & 34             & individual        & individual's behaviors \\ \hline
\end{tabular}%
}
\end{table*}

\subsection{Annotation}

\subsubsection{Person Bounding Box Annotation}In each video, four corners of every person appearing in the first frame are manually marked, and the smallest rectangular bounding box encompassing all four corners is computed and stored. Since multiple individuals typically appear in most videos, to better discern the order of each individual's appearance, we have assigned a color palette consisting of 30 different colors to the bounding boxes. This way, the sequence of appearances of different individuals within the same video is noted according to the color palette, and the sequence of individuals appearing in different videos employs the same color scheme. We have invested 300 man-hours in completing this phase of annotation.
\subsubsection{Captioning Individual Actions}The most labor-intensive aspect of our annotation process pertains to the captioning of individual actions. Given that numerous individuals appear within a single video, meticulous descriptions of each person's actions necessitate multiple video reviews, which is further complicated by the provision of bounding boxes for only the initial appearance frame of each individual. The captioning of each individual's actions commences from their first appearance-the frame marked with a bounding box and concludes when they vanish from the video. The content of each caption primarily consists of the scene in which the individual is located, their attire, a fine-grained objective description of their actions, and interactions with other individuals within the video. Generally, the length of a caption ranges between 15 to 45 words, but for extended videos, this can increase to approximately 65 words. Among all depicted actions, the most frequently occurring verbs include "walking", "looking", and "turning". Given the prevalence of anomalous behavior in the UCF-crime dataset, men are predominantly represented, with the most common scenes being "streets" and "indoors". All annotations are performed by trained native speakers.

\subsection{Comparison with Other Datasets}
Table~\ref{tab:dataset} compares our dataset, based on UCF-crime annotations and centered around human behavior, with other caption datasets. We compare UCCD with existing video caption datasets in several aspects: domain, video source, average time, caption length, caption target, and target type. As can be seen from the table, most videos are sourced from YouTube, while ours originate from real-world video surveillance . Due to our detailed, granular behavior descriptions of people, we have the longest caption length. Furthermore, the target type of other videos is generic events or actions, while ours is focused on providing detailed descriptions of individuals from multiple perspectives. Our dataset has certain unique features distinguishing it from other datasets, and these are summarized below:

\subsubsection{Data Source}

Unlike datasets like MSR-VTT, MSVD, ActivityNet Captions, etc., which are sourced from YouTube, our dataset is based on UCF-crime. Notably, our dataset includes a plethora of anomaly scene captions in addition to regular scenarios, setting it apart from conventional datasets.
\subsubsection{Video Integrity}Our dataset provides an extensive video surveillance  scene, detailing individual behaviors throughout the video with fine-grained descriptions. This is different from other video caption datasets that typically provide descriptions for short videos or carry out dense video captioning by breaking a video into several segments. Thus, our dataset boasts superior scene and temporal continuity.
\subsubsection{Behavioral Complexity}In contrast to video captions in normal scenarios like those in HowTo100M, YouCook2, and other existing video caption datasets, our dataset includes descriptions for various anomaly scenarios. These anomaly scenarios are known to entail more complex human behaviors, enhancing the richness and challenge of our dataset.
\subsubsection{Annotation Challenge}Throughout the captioning process, since only the bounding box of the first frame where a person appears is provided, the annotator must constantly track the individual and describe their interactions with others. This enhances the difficulty of annotation, but simultaneously improves the quality of the dataset.

In summary, our dataset excels in terms of data source, video integrity, behavioral complexity, and annotation challenge. This makes our dataset highly valuable to researchers.

\section{Approach}
In this section, we introduce a video captioning algorithm that takes advantage of the extensive annotations in our proposed dataset, capable of accurately describing the behavior of each individual appearing in the video. The details of our video captioning algorithm will be elaborated in the following content.

\begin{figure*}[!t]
\centering
\includegraphics[width=7in]{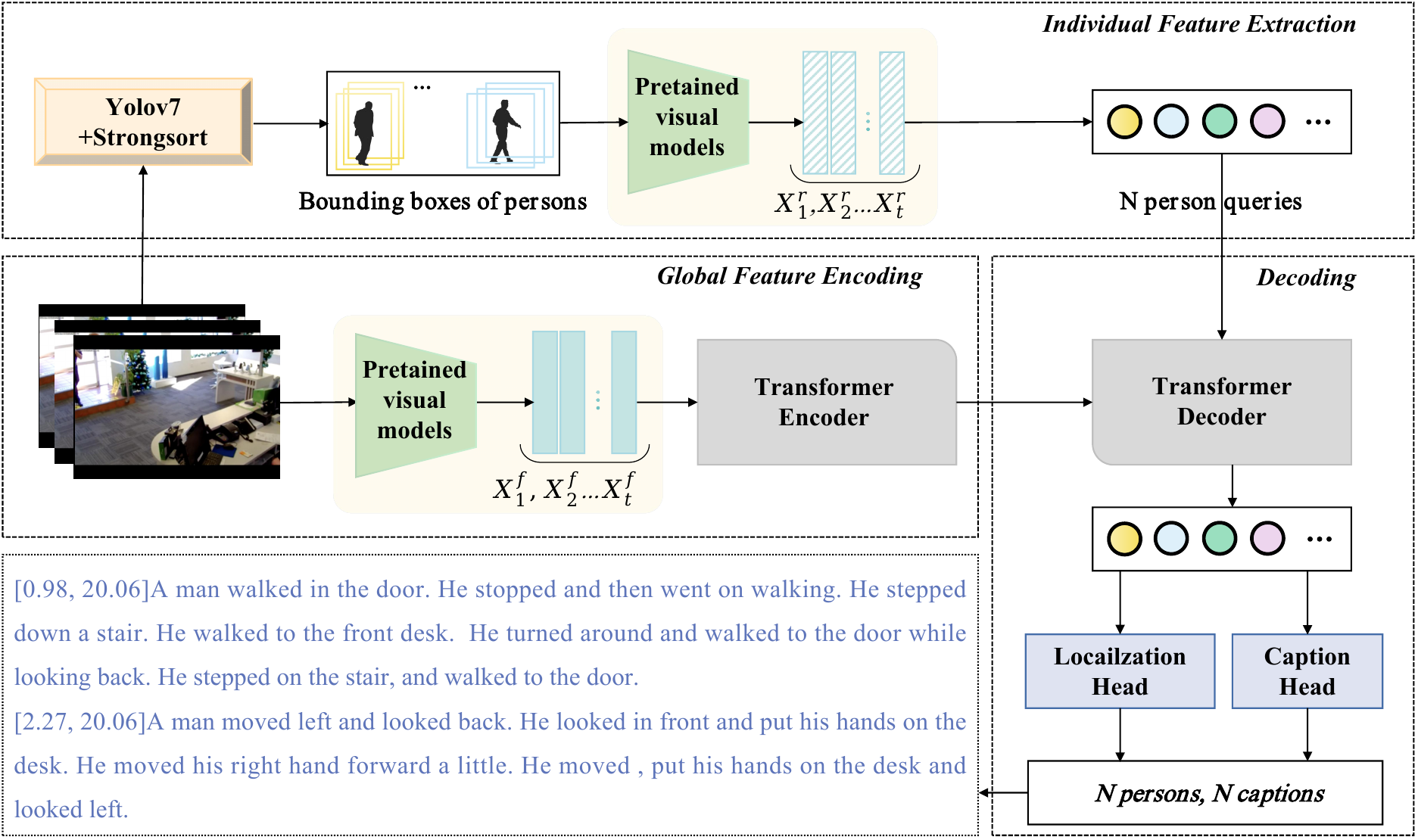}
\caption{The overall structure of our proposed model consists of three main components. The first component is feature extraction, in which we separately employ the pretrained visual models to extract global and individual features. The second component is feature processing, where we use a deformable transformer to concentrate attention on the times when individuals appear, thereby enhancing the performance of the model's captioning. The final component consists of a localization head and a captioning head, which generate the captions. The localization head is used to pinpoint the times when individuals appear and disappear, while the captioning head, comprised of elements such as an LSTM, is used to generate captions.}
\label{fig:framework}
\end{figure*}

\subsection{Overall Framework}
The introduced method, depicted in Figure~\ref{fig:framework}, consists of two primary components. The initial component deals with feature encoding; this process begins with frame extraction from the video, followed by the utilization of pretrained visual models~\cite{ref25,ref35,ref36} for frame-level feature extraction. The second stage involves feeding the extracted features, along with their respective positional embeddings~\cite{ref43}, into a deformable encoder~\cite{ref42}. Then the Yolov7~\cite{ref83} + Strongsort~\cite{ref84} with OsNet~\cite{ref44} is used to detect and track individuals in the video, cropping the video during the time each individual appears and disappears according to the bounding boxes and extracting frames. The frames that contain only individual's information are used to extract features with the same pretrained visual models~\cite{ref25,ref35,ref36}, and the obtained features of each individual are input into the deformable decoder~\cite{ref42} as a query.

The second component involves decoding, where the features of the individuals in the video are extracted and combined with the frame features in the encoder, before being placed into the decoder. The decoder outputs the queried features~\cite{ref55} connected to a localization head and a captioning head for generating each person's caption. The loss function includes localization loss and captioning loss, used respectively for calculating the timing of the appearance and disappearance of individuals, and for comparing the generated captions with the real captions.

\subsection{Feature encoding}
Our initial step towards exploiting the comprehensive spatio-temporal characteristics within a video is to utilize pretrained visual models to perform feature extraction at the frame level. We employ a consistent frame rate of 30 fps to uniformly sample frames from the video. Each frame, represented as $x_{img} \in  R^{3\times H_0\times W_0}$ , is processed through I3D to extract features represented as $f \in R^{C\times H\times W}$, with our conventional values being $C = 1024$ and $H,W = H_0/32, W_0/32$.

We handle the video as a series of frames, and for a temporal sequence length of $t$, we achieve a set of frame features denoted as $X_1^f$, $X_2^f$,..., $X_t^f$, where $X_i^f$ denotes the feature vector for the $i^{th}$ frame. Following this, we employ a $1 \times 1$ convolution to reduce the channel dimension from $C$ to a smaller dimension d in the high-level activation map $f$, generating a new feature map $z_0 \in R^{d\times H\times W}$. Given that the encoder expects a sequence as input, we collapse the spatial dimensions of $z_0$ into one dimension, forming a feature map of $d\times HW$. Each layer within the encoder is designed with a standard architecture, composed of a multi-head self-attention module and a feed-forward network. Since the transformer architecture is permutation-invariant, it is supplemented with fixed positional encodings which are introduced to the input of each attention layer.


\subsection{Person Feature Extraction}
Our approach diverges from traditional video captioning methods, which typically focus on depicting specific events. Instead, our strategy centers on comprehensively narrating all actions performed by participants in the video, requiring exhaustive feature extraction at the individual level. During the detection phase, we use Yolov7+Strongsort with OsNet~\cite{ref44} for detecting and tracking individuals. This algorithm incorporates reid~\cite{ref61} capabilities, addressing issues such as losing original identification recognition of the individual after individuals meet or reappear after a period of absence. We obtain frames labeled with individual bounding boxes, denoted as $F_i$. Subsequently, we segregate regions containing these bounding boxes, yielding the set $R_i$. In order to more accurately capture the actions of each participant during the interaction process via features, we employ a frame feature extraction technique to identify distinct attributes of each individual. We apply uniform frame sampling, extracting 64 frames from each person, with an input size of $224\times 224$, and then input them into I3D to obtain individual features, ultimately generating the feature set $X_1^r$, $X_2^r$,..., $X_t^r$. Upon completing the pooling process, we obtain an output dimension of $1024\times t$. These extracted features are then passed through a fully connected(FC) layer to be converted into a compact 256-dimensional format. Each query stands for a single individual's features, and then these $N$ distinct queries are fed into the Transformer decoder's multi-head self-attention layer.

\subsection{Decoding}
The decoding network comprises three main components: a Deformable Transformer Decoder, and two parallel heads - a captioning head for generating captions, and a localization head designed to predict human boundaries. The Deformable Transformer is an encoder-decoder architecture based on multi-scale deformable attention (MSDAtt), which mitigates the slow convergence problem of the self-attention in Transformer when processing image feature maps, by attending to a sparse set of sampling points around reference points. Given multi-scale feature maps $X $, where $X \in R^{C\times H\times W} $, a query element $q_j$ and a normalized reference point $p_j \in [0, 1]^2$, MSDAtt outputs a context vector~\cite{ref56} by the weighted sum of $K$ sampling points across feature maps at $L$ scales.

The goal of the decoder is to query frame-level features of human features under the condition of N human features ${q_j}_{j=1}^N$ and their corresponding scalar reference points pj. It is worth noting that ${q_j}$ is predicted by linear projection $p_j$ and using a Sigmoid activation function. Human features and reference points serve as initial guesses for human features and positions, and they interact with each other at each decoding layer. The output query features and reference points are denoted as $q_j$ and $p_j$, respectively.

To distinguish characters with overlapping characteristics, a localization head is trained, which performs box prediction and binary classification for each unique character feature. Box prediction aims to predict the 2D relative offset of ground truth segments, corresponding to specific reference points. The goal of binary classification is to generate foreground confidence scores for each character query. The mechanisms for box prediction and binary classification are both facilitated by a multilayer perceptron. As such, a set of tuples ${\{t_j^s, t_j^e, c_j^{loc}}\}_{j=1}^N$ are obtained, representing the start time, end time, and location of the detected characters. Here, $c_j^{loc}$ represents the location confidence of character query $\Tilde {q_j}$.

\begin {align} 
{MSDAtt} ({q}_j,  {p}_j,  {X}) = \sum _{l=1}^L \sum _{k=1}^K A_{jlk}  {W}  {x}^l_{\Tilde {{p}}_{jlk}} \label {eq2}
\end {align} 

\begin {align}
\Tilde { {p}}_{jlk} = \phi _l( {p}_j) + \Delta {p}_{jkl}\label {eq3} 
\end {align}

For creating descriptive content for video captions, a different task setup than traditional methods~\cite{ref58} is adopted. A new method is proposed, which uses LSTM hidden state $h_{jt}$ to predict the word $w_{jt}$ after applying a fully connected layer and softmax activation, instead of inputting character-level representation $q_j$ into a standard LSTM at each timestamp. The standard captioning model, considering only character-level representation $q_j$ lacks dynamic interaction between linguistic cues and frame features. To rectify this, a mechanism based on soft attention, known as Deformable Soft Attention (DSA), is introduced. This mechanism effectively enforces soft attention weights to concentrate in a smaller region around the reference point. When generating the t-th word wt, $K$ sampling points are first created on each $f^l$ using linguistic query $h_{jt}$ and character query $q_j$ , following Equation~\ref{eq2}, where $h_{jt}$ represents the hidden state within the LSTM. Then, the $K$ sampling points are considered as key values, and [$h_{jt}$, $q_j$] are considered as the query inside the soft attention. Considering that sampling points are distributed around reference point $p_j$, the output feature $z_{jt}$ of DSA is constrained within a relatively small region. LSTM takes concatenated context features $z_{jt}$, character query features $q_j$ , and previous word ${\{w_j,t-1\}}$ as input. By applying softmax activation to $h_{jt}$, the probability of the subsequent word $w_{jt}$ is obtained. As LSTM proceeds, a sentence $S_j = w_{j1}, ..., w_{jMj}$ is generated, where $M_j$ indicates the length of the sentence.

\subsection{Loss Function}
In the course of training, our model generates a set of actions for $N$ individuals, encompassing both location and description. To align predicted events with actual data in the global scheme, we employ the Hungarian algorithm as per~\cite{ref47} to determine the optimal binary matching outcome. The matching cost is defined as $C = \alpha _{giou}\mathcal{L}_{giou} + \alpha _{cls}\mathcal{L}_{cls}$, where $\mathcal{L}_{giou}$ indicates the generalized IOU~\cite{ref49} between predicted and actual time segments, while $\mathcal{L}_{cls}$ refers to the focal loss~\cite{ref48} between the predicted classification score and actual data labels. The chosen pairs are used to calculate the set prediction loss, which is a weighted sum of gIOU loss, classification loss and caption loss: 

\begin{align}
\mathcal{L} = \beta_{giou}\mathcal{L}_{giou} + \beta_{cls}\mathcal{L}_{cls} + \beta_{cap}\mathcal{L}_{cap}
\end{align}

Here, $\mathcal{L}_{cap}$ measures the cross-entropy between predicted word probabilities and actual values, normalized by caption length. The $\beta$ represents the weights of various losses, such as gIOU loss, classification loss, and caption loss. Importantly, we adhere to~\cite{ref51,ref47} in adding a prediction head at every layer of the transformation decoder. The final loss is the sum of the set prediction losses across all decoder layers.

\section{Experiments}

In this section, we present the trial outcomes of our proposed video captioning technique on our UCCD. This includes ablation studies, comparative trials against both baseline methods and the latest advancements in video captioning techniques. Additionally, we conduct an evaluation of human performance to gauge the potential of our dataset. For the sake of clarity, we initially provide the evaluation procedures and details of implementation.

\begin{figure*}
\centering
\includegraphics[width=7in]{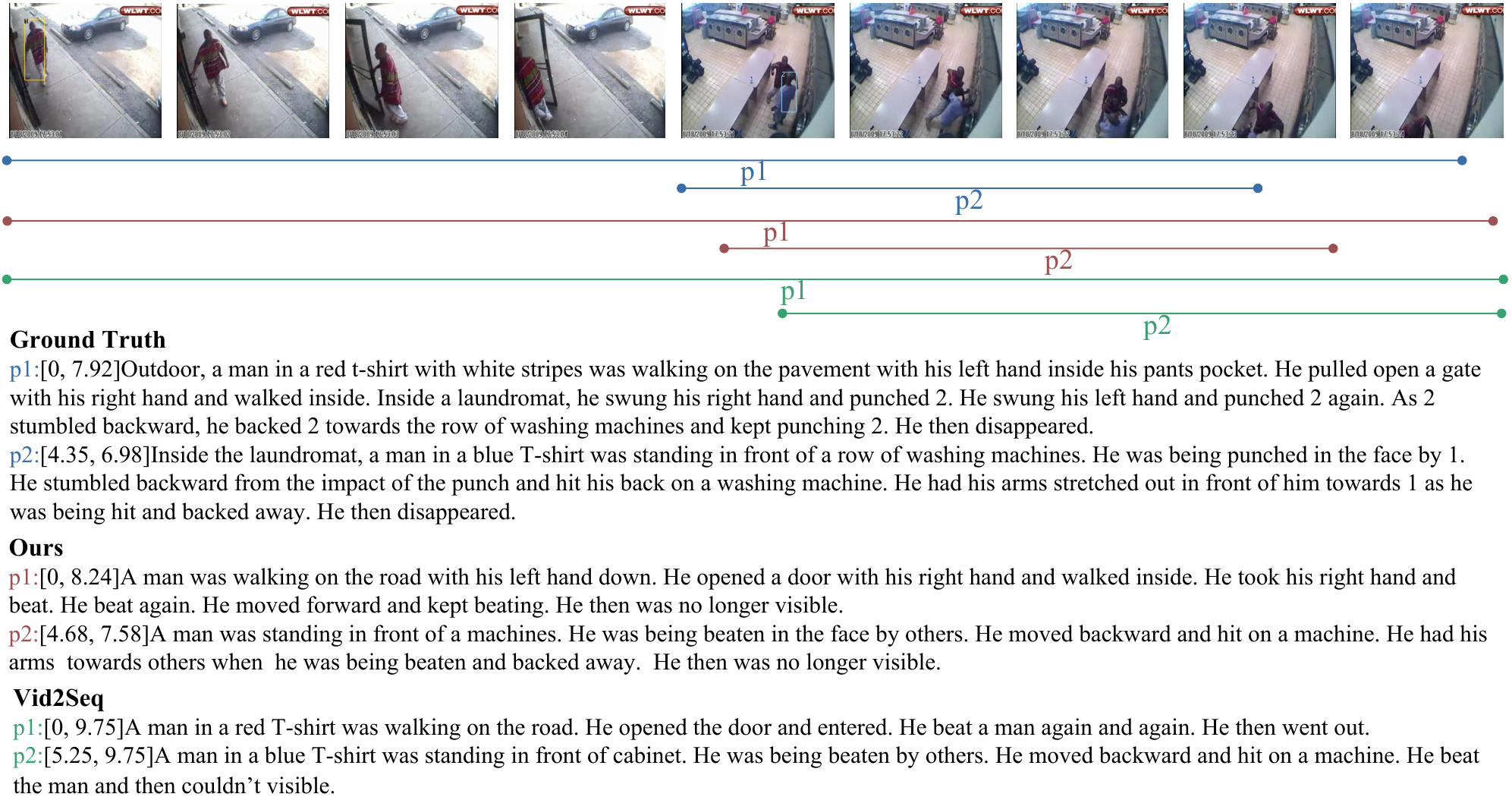}
\caption{The figure shows the visualization results of video captioning. $p_1$ and $p_2$ represent two individuals generating two captions. Different colors indicate different models. The horizontal line above represents the timeline, which allows for a more intuitive observation of the model's localization performance.}
\label{fig:sample}
\end{figure*}

\subsection{Evaluation Protocols and Implementation Details}
The UCCD is randomly divided into training, validation, and testing subsets. The training subset encompasses 584 videos with 4014 captions, the validation subset contains 205 videos and 1842 captions, and the testing subset includes 223 videos with 1964 captions. In addition to the captions, each video has time locations corresponding to the number of captions, denoting the time the person first appears in the video in each caption. Our method's captioning performance is evaluated using BLEU4~\cite{ref20}, METEOR~\cite{ref21}, CIDER~\cite{ref23}, and ROUGE-L~\cite{ref22}, calculating the average precision of matches between the generated captions and the benchmark true captions at IOU thresholds {0.3, 0.5, 0.7, 0.9}. However, these scorers do not consider narrative quality or how well the generated captions cover the entire video story. Hence, we further employ SODA\_c~\cite{ref24} for comprehensive evaluation. During the tracking and detection phase, we experiment with the yolov5+deepsort and Yolov7+Strongsort with OsNet methods. For UCF-crime caption, we tested our model using three distinct feature extraction methods: C3D~\cite{ref25}, I3D~\cite{ref35}, and CLIP~\cite{ref36}. We used the C3D pre-trained on Sports1M~\cite{ref26} to extract frame-level features. For I3D, we used a model pre-trained on the Kinetics dataset~\cite{ref35}. For CLIP, we downloaded a pre-trained model released officially by OpenAI. We used a two-layer deformable transformer with multi-scale (4-level) deformable attention. The deformable transformer uses a hidden size of 512 in the MSDAtt layer and 2,048 in the feed-forward layer. After each person is detected and tracked, their features are separately extracted using the corresponding feature extraction method and fed as a query into the deformable transformer. Finally, we employed an LSTM as the caption generator with a hidden dimension of 512. We use the Adam optimizer~\cite{ref27} with an initial learning rate of 5e-5, with each mini-batch sized to one video. Two NVIDIA A100 GPUs facilitate the training process.

\subsection{Comparison with State-of-the-art Methods}

\noindent \textbf{Overall Introduction.} Although specialized methods for annotating individual actions across diverse scenarios are limited, we are dedicated to addressing this issue. In our approach, we utilize the I3D model pre-trained on the Kinetics dataset to extract human behavioral features, as Kinetics is a large-scale human action video dataset that is more suitable for our task. We employ Yolov7+Strongsort and OSnet to detect and track each individual's features~\cite{ref57}, using I3D to extract them, and interact with the overall conditional features to enhance the quality of individual subtitle generation~\cite{ref58}. As shown in Figure~\ref{fig:sample}, we have visualized the subtitle results outputted by our model and compared them with existing methods.

\noindent \textbf{Method Comparison.} We compared our algorithm with nine leading technologies applied to the most commonly used datasets across different scenarios, such as ActivityNet Captions, MSR-VTT, and VATEX. The results, as shown in Table~\ref{tab:compare}, indicate that our algorithm outperforms the other nine state-of-the-art methods, demonstrating our algorithm's outstanding performance. In terms of BLEU-4, CIDER, METEOR, ROUGE-L metrics, we have respectively achieved improvements of 4.2, 3.8, 3.6, and 1.3 over the current sota, highlighting our method's advantages in character positioning and action description.

\begin{table*}[htbp]
\centering
\caption{Compare our model with other baselines and various intermediate models.We can find that I3D has greater contribution for improving performance}
\label{tab:compare}
\setlength{\tabcolsep}{4.6mm}
\begin{tabular}{l|c|cccc|c}
\hline
Method        & Features     & BLEU-4        & CIDER         & METEOR        & ROUGE-L       & Extra Data Training                 \\ \hline
Vid2Seq~\cite{yang2023vid2seq}         & CLIP ViT-L/14            & 40.5         & 71.4          & 26.7          & 57.9          & \checkmark           \\
VALOR~\cite{ref28}         & CLIP/VideoSwin Transformer            & 34.6          & 62.3          & 19.4          & 48.7          & \checkmark           \\
mPlug-2~\cite{ref30}       & Dual-vision Encoder            & 31.7          & 61.4          & 16.2          & 43.6          & \ding{55}           \\
VAST~\cite{ref31}          &  Vision Transformer             & 34.1          & 64.2          & -             & -             & \checkmark           \\
GIT2~\cite{ref32}          & Florence            & 29.6          & 57.8          & 15.2          & 41.1          & \checkmark           \\
VLAB~\cite{ref33}          & CLIP            & 29.4          & 56.7          & 14.5          & 41.3          & \checkmark           \\ \hline
MT~\cite{zhou2018end}            & TSN          & 32.8          & 60.2          & 16.3          & 44.5          & \ding{55}          \\
BMT~\cite{iashin2020better}           & I3D          & 34.3          & 62.7          & 18.6          & 46.8          & \ding{55}         \\
PDVC~\cite{ref17}          & TSP          & 36.5          & 65.4          & 20.7          & 49.7          & \ding{55}          \\ \hline
Ours          & C3D          & 43.6          & 73.8          & 28.5          & 57.1          & \ding{55}           \\
Ours          & CLIP         & 44.2          & 74.6          & 29.2          & 57.9          & \ding{55}           \\
\textbf{Ours} & \textbf{I3D} & \textbf{44.7} & \textbf{75.2} & \textbf{30.3} & \textbf{59.2} & \textbf{\ding{55}}   \\ \hline
\end{tabular}%
\end{table*}

\noindent \textbf{Reason Analysis.} Currently, the cutting-edge technology for dense video subtitling is Vid2Seq~\cite{yang2023vid2seq}, and the most advanced methods for the MSR-VTT dataset include mPlug-2~\cite{ref30}, VAST~\cite{ref31}, GIT2~\cite{ref32}, VLAB~\cite{ref33}, and VALOR~\cite{ref28}, with VALOR and VAST leading in the VATEX dataset. Next, we analyze the reasons why these methods do not perform well on the UCCD dataset. Vid2Seq~\cite{yang2023vid2seq} enhances the language model through special time marking but performs poorly in fine-grained description. mPlug-2~\cite{ref30} resolves the entanglement between video-text modalities but falls short in tracking individuals. VAST~\cite{ref31} achieves state-of-the-art results in video subtitling tasks but struggles to capture actions effectively. Similarly, GIT2~\cite{ref32} and VALOR~\cite{ref28} demonstrate strong generalization abilities in pre-training but face difficulties in extracting individual actions. We also employed counters from MT~\cite{zhou2018end}, BMT~\cite{iashin2020better}, PDVC~\cite{ref17} to distinguish different individuals in our task, and the experimental results showed that their effects were not as good as the detection and tracking of people in our methods. Therefore, the performance of these methods on the UCCD dataset is not very satisfactory.


\begin{table*}
\centering
\caption{Experimental results for video captioning in different scenarios. In both normal and anomaly scenarios, our method achieved the best results across all metrics}
\label{tab:normal}
\setlength{\tabcolsep}{4.5mm}
{%
\begin{tabular}{l|cccc|cccc}
\hline
\multirow{2}{*}{Method} & \multicolumn{4}{c|}{Normal Scenarious} & \multicolumn{4}{c}{Anomaly Scenarious} \\
                  & BLEU-4   & CIDER  & METEOR  & ROUGE-L  & BLEU-4   & CIDER   & METEOR  & ROUGE-L  \\ \hline
VALOR~\cite{ref28}             & 34.6     & 62.3   & 19.4    & 48.7     & 33.6     & 59.3    & 16.2    & 47.7     \\
VAST~\cite{ref31}              & 34.1     & 64.2   & -       & -        & 32.1     & 60.3    & -       & -        \\
OURS              & 44.7     & 75.2   & 30.3    & 59.2     & 44.2     & 74.8    & 30.6    & 58.7     \\ \hline
\end{tabular}%
}
\end{table*}


\begin{table*}
\centering
\caption{Results of the Model Using Different Person Tracking Detection Algorithms. Under the condition of using the same feature extraction method, the experimental results of YOLOv7 are the best, indicating that it is more suitable for our task}
\label{tab:track}
\setlength{\tabcolsep}{9.7mm}
{%
\begin{tabular}{l|c|c|c|c|c}
\hline
{\color[HTML]{000000} Method} & Features & {\color[HTML]{000000} BLEU-4} & {\color[HTML]{000000} CIDER} & {\color[HTML]{000000} METEOR} & {\color[HTML]{000000} ROUGE-L} \\ \hline
{\color[HTML]{000000} BBOX}   & C3D      & 42.1                          & 73.4                         & 26.3                          & 56.2                           \\
{\color[HTML]{000000} POS}    & C3D      & 42.8                          & 73.3                         & 27.6                          & 56.8                           \\
{\color[HTML]{000000} YOLOV5} & C3D      & 38.5                          & 69.3                         & 21.6                          & 50.4                           \\
{\color[HTML]{000000} YOLOV7} & C3D      & 43.6                          & 73.8                         & 28.5                          & 57.1                           \\ \hline
\end{tabular}%
}
\end{table*}

\subsection{Anomalies Captioning}

Regarding the peculiarities of our dataset, it is worth noting that it is annotated based on the UCF-crime dataset, which is divided into normal videos and videos showcasing anomaly scenarios. These comprise 13 types of real-life anomalies, namely, abuse, arrests, arson, assaults, road accidents, burglary, explosions, fights, robbery, shooting, theft, shoplifting, and vandalism. These particular anomaly behaviors were chosen due to their detrimental impact on public safety. To validate our method's capacity to generate captions for human behavior under anomaly scenarios, we proceeded to re-segment the UCCD. We designated all normal videos to the training set, amassing a total of 630 videos, while all anomaly scenarios were allocated to the validation set, accounting for 382 videos. As illustrated in Table~\ref{tab:normal}, from our experimental results, in normal scenarios, the various metrics have been improved by 10.1, 11.0, 10.9, and 10.5, respectively, compared to the state-of-the-art methods benchmarked on the VATEAX dataset. Even under anomaly scenarios, the metrics of our method have been increased by 10.6, 14.5, 14.4, and 11.0, respectively, achieving commendable results.

\subsection{Person Tracking Detection}
For the experiment of person tracking detection, we utilize the pre-trained models Yolov5+Deepsort and Yolov7+Strongsort with OsNet on the COCO dataset to perform human detection and tracking. Given that Yolov5 and Yolov7 are designed for multi-type object detection, whereas tracking systems can only track one type of object, we limit the number of detection types to a single class for tracking purposes, i.e., classes 1 for humans.

Upon detecting and tracking the presence of humans, the content within each person's bounding box is cropped and saved as a new video. The original frames are cropped based on each individual's bounding box and then saved directly to the corresponding videowriter. As each person's bounding box size varies, the resolution of each created video differs as well. To maintain consistency in feature extraction, the methodology of extracting individual features remains the same as that for global feature extraction.

Based on the different feature extraction techniques employed subsequently, the videos are resized to the corresponding resolution. For instance, with C3D for feature extraction, the resolution is set at $112\times112$, with a frame rate of 30fps, and every 16 frames are selected as a person's feature. When using I3D for feature extraction, the resolution is $224\times224$, with a frame rate of 30fps, and every 64 frames are selected as a person's feature. When using CLIP for feature extraction, the resolution is $224\times224$, with a frame rate of 30fps, and every 40 frames are chosen as a person's feature.

Apart from these methods, we also tried two other techniques. The first method involves cropping the content in the bounding box of the first frame in which each person appears and then performing feature extraction using the corresponding feature extraction method. The second method involves encoding all the locations where a person has been tracked into the model as a query. Table~\ref{tab:track} displays the impact of different person tracking detection algorithms on model performance.


\begin{table*}[htbp]
\centering
\caption{Ablation studies on the UCF-crime captioning dataset validation set. We test the components' impact on our experimental results from four aspects: Tracking Detection, Transformer, Localization Head, and Captioning Head}
\label{tab:Ablation studies}
\setlength{\tabcolsep}{3.7mm}
{%
\begin{tabular}{cc|cc|cc|ccc|cc}
\hline
\multicolumn{2}{c|}{Tracking Detection} & \multicolumn{2}{c|}{Transformer} & \multicolumn{2}{c|}{Localization Head} & \multicolumn{3}{c|}{Captioning Head} & \multirow{2}{*}{METEOR} & \multirow{2}{*}{SODA\_c} \\
Yolov7+Strongsort         & w/o         & Vanilla       & Deformable       & w                 & w/o                & LSTM        & SA        & DSA        &                         &                         \\ \hline
                          & \checkmark           &               & \checkmark                & \checkmark                 &                    & \checkmark           &           & \checkmark          & 26.4                    & 20.3                    \\
\checkmark                         &             & \checkmark             &                  & \checkmark                 &                    & \checkmark           &           &            & 28.7                    & 22.8                    \\
\checkmark                         &             &               & \checkmark                &                   & \checkmark                  & \checkmark           &           & \checkmark          & 29.8                    & 26.4                    \\
\checkmark                         &             &               & \checkmark                & \checkmark                 &                    & \checkmark           &           &            & 29.6                    & 26.1                    \\
\checkmark                         &             &               & \checkmark                & \checkmark                 &                    & \checkmark           & \checkmark         &            & 29.3                    & 23.2                    \\
\checkmark                         &             &               & \checkmark                & \checkmark                 &                    & \checkmark           &           & \checkmark          & \textbf{30.3}           & \textbf{26.8}           \\ \hline
\end{tabular}%
}
\end{table*}

\subsection{Ablation Study}
This section aims to assess the efficiency of the proposed approach and showcase the contributions of each component of our suggested model towards the final performance. We depict the performance of action captioning by comparing it with four influential ablation studies. For assessment, the generated caption is evaluated using METEOR and SODA\_c metrics, as detailed in Table~\ref{tab:Ablation studies}.

\subsubsection{Tracking Detection}
Our first experiment revolves around the examination of the influence of person detection and tracking algorithms on the model's performance. The results suggest that engaging in detection, tracking, and feature extraction for each individual, coupled with an interaction with the global features, greatly affects the model's overall performance.

\subsubsection{Transformer}
Transitioning from a deformable transformer to a vanilla one also impacts the model's performance. We observed that incorporating locality into the transformer effectively aids in the extraction of temporally-sensitive features, which is crucial for localization-aware tasks.

\subsubsection{Localization Head}
We further add a localization head in our model, which has shown to improve the determination of the start and end times of events. In other words, it accurately tracks the appearance and disappearance times of each individual.

\subsubsection{Caption Head}
Regarding caption generation, LSTM is primarily employed, but we have enhanced it by adding two distinct attention mechanisms. This addition helps LSTM in generating more accurate descriptions of human behavior. Our results clearly show that focusing on a small segment around the proposals, instead of the entire video, significantly optimizes behavior captioning.


\subsection{Human Performance Evaluation}
In order to explore the complexity of our dataset and the performance difference between human evaluators and our algorithm, we conducted a human performance assessment on our dataset. In this experiment, we randomly selected 10 videos under anomaly scenarios, each video containing an average of about 5 individuals, each appearing for a duration varying from 5 to 30 seconds. Three well-trained annotators participated in this experiment, and their performances are displayed in Table~\ref{tab:Human}. According to the data from the table, the accuracy rate of human annotators is on average 15\% higher than that of our model. Despite human annotators providing precise descriptions, the time cost is significantly high. Annotating each video requires an average of 13 minutes per video, while our trained model can generate captions within a few seconds.

\begin{table}
\centering
\caption{Comparison between human evaluators and our method. Our findings reveal that the average accuracy of human assessments stands at 91\%, outperforming our algorithm by a significant 15\%. this sizable gap underscores the substantial potential for further enhancing our algorithm's performance} 
\label{tab:Human}
\resizebox{\columnwidth}{!}{%
\begin{tabular}{c|cccc}
\hline
         & Annotator 1 & Annotator 2 & Annotator 3 & Ours \\ \hline
Accuracy & 92\%        & 90\%        & 91\%        & 76\% \\ \hline
\end{tabular}%
}
\end{table}

\section{Conclusion}
In this paper, we have gathered the UCF-crime captioning dataset, the first of its kind, to our knowledge, that offers captions for anomalous videos in real-life surveillance scenarios. We furnish the bounding box for the first frame each individual appears in and proceed to provide an objective description of the person's entire behavior in the video. Consequently, our dataset can also be applied to other vision tasks, such as action recognition and anomalyity detection in videos. Moreover, during the captioning of individuals, we have also included some information about their interactions with other people; how to capture this information is worthy of further exploration. In addition, we have conducted thorough experiments to fully exploit the rich annotations. Drawing on the abundant annotation data in our dataset, we have proposed a novel method for captioning people's behaviors in videos. This method can detect and track individuals at every time point in the video and provide an objective description of their behaviors. Experimental results show that our proposed method significantly outperforms five state-of-the-art video captioning methods evaluated on our dataset.


\vfill


\begin{thebibliography}{1}
\bibliographystyle{IEEEtran}

\bibitem{ref45}Shidik, G., Noersasongko, E., Nugraha, A., Andono, P., Jumanto, J. \& Kusuma, E. A systematic review of intelligence video surveillance: trends, techniques, frameworks, and datasets. {\em IEEE Access}. \textbf{7} pp. 170457-170473 (2019)


\bibitem{ref38}Song, H., Sun, C., Wu, X., Chen, M. \& Jia, Y. Learning Normal Patterns via Adversarial Attention-Based Autoencoder for Abnormal Event Detection in Videos. {\em IEEE Transactions On Multimedia}. \textbf{22}, 2138-2148 (2020)


\bibitem{ref39}Morais, R., Le, V., Tran, T., Saha, B., Mansour, M. \& Venkatesh, S. Learning regularity in skeleton trajectories for anomaly detection in videos. {\em Proceedings Of The IEEE/CVF Conference On Computer Vision And Pattern Recognition}. pp. 11996-12004 (2019)


\bibitem{ref40}Li, N., Chang, F. \& Liu, C. Spatial-Temporal Cascade Autoencoder for Video Anomaly Detection in Crowded Scenes. {\em IEEE Transactions On Multimedia}. \textbf{23} pp. 203-215 (2021)

\bibitem{ref41}Sultani, W., Chen, C. \& Shah, M. Real-world anomaly detection in surveillance videos. {\em Proceedings Of The IEEE Conference On Computer Vision And Pattern Recognition}. pp. 6479-6488 (2018)

\bibitem{ref62}Wu, P., Liu, J. \& Shen, F. A deep one-class neural network for anomalous event detection in complex scenes. {\em IEEE Transactions On Neural Networks And Learning Systems}. \textbf{31}, 2609-2622 (2019)

\bibitem{ref63}Wu, P., Liu, J., Shi, Y., Sun, Y., Shao, F., Wu, Z. \& Yang, Z. Not only look, but also listen: Learning multimodal violence detection under weak supervision. {\em Computer Vision–ECCV 2020: 16th European Conference, Glasgow, UK, August 23–28, 2020, Proceedings, Part XXX 16}. pp. 322-339 (2020)

\bibitem{ref64}Wu, P. \& Liu, J. Learning causal temporal relation and feature discrimination for anomaly detection. {\em IEEE Transactions On Image Processing}. \textbf{30} pp. 3513-3527 (2021)

\bibitem{ref65}Yang, Z., Liu, J., Wu, Z., Wu, P. \& Liu, X. Video Event Restoration Based on Keyframes for Video Anomaly Detection. {\em Proceedings Of The IEEE/CVF Conference On Computer Vision And Pattern Recognition}. pp. 14592-14601 (2023)

\bibitem{ref44}Quan, H. \& Ablameyko, S. Multi-object tracking by using strong SORT tracker and YOLOv7 network. (2022)

\bibitem{ref52}Bi, Y., Jiang, H., Hu, Y., Sun, Y. \& Yin, B. See and Learn More: Dense caption-aware Representation for Visual Question Answering. {\em IEEE Transactions On Circuits And Systems For Video Technology}. pp. 1-1 (2023)





\bibitem{ref1}Xu, J., Mei, T., Yao, T. \& Rui, Y. Msr-vtt: A large video description dataset for bridging video and language. {\em Proceedings Of The IEEE Conference On Computer Vision And Pattern Recognition}. pp. 5288-5296 (2016)

\bibitem{ref2}Miech, A., Zhukov, D., Alayrac, J., Tapaswi, M., Laptev, I. \& Sivic, J. Howto100m: Learning a text-video embedding by watching hundred million narrated video clips. {\em Proceedings Of The IEEE/CVF International Conference On Computer Vision}. pp. 2630-2640 (2019)

\bibitem{ref3}Krishna, R., Hata, K., Ren, F., Fei-Fei, L. \& Carlos Niebles, J. Dense-captioning events in videos. {\em Proceedings Of The IEEE International Conference On Computer Vision}. pp. 706-715 (2017)

\bibitem{ref50}Wang, Y., Gao, D., Yu, L., Lei, W., Feiszli, M. \& Shou, M. GEB+: A Benchmark for Generic Event Boundary Captioning, Grounding and Retrieval. {\em Computer Vision–ECCV 2022: 17th European Conference, Tel Aviv, Israel, October 23–27, 2022, Proceedings, Part XXXV}. pp. 709-725 (2022)

\bibitem{ref4}Zhou, L., Xu, C. \& Corso, J. Towards automatic learning of procedures from web instructional videos. {\em Proceedings Of The AAAI Conference On Artificial Intelligence}. \textbf{32} (2018)

\bibitem{ref5}Chen, D. \& Dolan, W. Collecting highly parallel data for paraphrase evaluation. {\em Proceedings Of The 49th Annual Meeting Of The Association For Computational Linguistics: Human Language Technologies}. pp. 190-200 (2011)


\bibitem{ref6}Bain, M., Nagrani, A., Varol, G. \& Zisserman, A. Frozen in time: A joint video and image encoder for end-to-end retrieval. {\em Proceedings Of The IEEE/CVF International Conference On Computer Vision}. pp. 1728-1738 (2021)

\bibitem{ref7}Wang, X., Wu, J., Chen, J., Li, L., Wang, Y. \& Wang, W. Vatex: A large-scale, high-quality multilingual dataset for video-and-language research. {\em Proceedings Of The IEEE/CVF International Conference On Computer Vision}. pp. 4581-4591 (2019)

\bibitem{ref8}Li, Y., Song, Y., Cao, L., Tetreault, J., Goldberg, L., Jaimes, A. \& Luo, J. TGIF: A new dataset and benchmark on animated GIF description. {\em Proceedings Of The IEEE Conference On Computer Vision And Pattern Recognition}. pp. 4641-4650 (2016)

\bibitem{ref9}Li, L., Lei, J., Gan, Z., Yu, L., Chen, Y., Pillai, R., Cheng, Y., Zhou, L., Wang, X., Wang, W. \& Others Value: A multi-task benchmark for video-and-language understanding evaluation. {\em ArXiv Preprint ArXiv:2106.04632}. (2021)


\bibitem{ref10}Li, L., Chen, Y., Cheng, Y., Gan, Z., Yu, L. \& Liu, J. Hero: Hierarchical encoder for video+ language omni-representation pre-training. {\em ArXiv Preprint ArXiv:2005.00200}. (2020)

\bibitem{ref11}Parmar, P. \& Morris, B. What and how well you performed? a multitask learning approach to action quality assessment. {\em Proceedings Of The IEEE/CVF Conference On Computer Vision And Pattern Recognition}. pp. 304-313 (2019)

\bibitem{ref12}Lei, J., Yu, L., Berg, T. \& Bansal, M. Tvr: A large-scale dataset for video-subtitle moment retrieval. {\em Computer Vision–ECCV 2020: 16th European Conference, Glasgow, UK, August 23–28, 2020, Proceedings, Part XXI 16}. pp. 447-463 (2020)

\bibitem{ref28}Chen, S., He, X., Guo, L., Zhu, X., Wang, W., Tang, J. \& Liu, J. Valor: Vision-audio-language omni-perception pretraining model and dataset. {\em ArXiv Preprint ArXiv:2304.08345}. (2023)

\bibitem{ref29}Gemmeke, J., Ellis, D., Freedman, D., Jansen, A., Lawrence, W., Moore, R., Plakal, M. \& Ritter, M. Audio set: An ontology and human-labeled dataset for audio events. {\em 2017 IEEE International Conference On Acoustics, Speech And Signal Processing (ICASSP)}. pp. 776-780 (2017)

\bibitem{ref13}Krishna, R., Hata, K., Ren, F., Fei-Fei, L. \& Carlos Niebles, J. Dense-captioning events in videos. {\em Proceedings Of The IEEE International Conference On Computer Vision}. pp. 706-715 (2017)


\bibitem{ref60}Liu, Y., Li, H., Cheng, J. \& Chen, X. MSCAF-Net: A General Framework for Camouflaged Object Detection via Learning Multi-Scale Context-Aware Features. {\em IEEE Transactions On Circuits And Systems For Video Technology}. pp. 1-1 (2023)


\bibitem{ref14}Wang, J., Jiang, W., Ma, L., Liu, W. \& Xu, Y. Bidirectional attentive fusion with context gating for dense video captioning. {\em Proceedings Of The IEEE Conference On Computer Vision And Pattern Recognition}. pp. 7190-7198 (2018)


\bibitem{ref15}Yang, D. \& Yuan, C. Hierarchical context encoding for events captioning in videos. {\em 2018 25th IEEE International Conference On Image Processing (ICIP)}. pp. 1288-1292 (2018)

\bibitem{ref18}Ghaderi, Z., Salewski, L. \& Lensch, H. Diverse Video Captioning by Adaptive Spatio-temporal Attention. {\em Pattern Recognition: 44th DAGM German Conference, DAGM GCPR 2022, Konstanz, Germany, September 27–30, 2022, Proceedings}. pp. 409-425 (2022)

\bibitem{ref19}Wang, L., Shang, C., Qiu, H., Zhao, T., Qiu, B. \& Li, H. Multi-stage tag guidance network in video caption. {\em Proceedings Of The 28th ACM International Conference On Multimedia}. pp. 4610-4614 (2020)

\bibitem{ref16}Li, Y., Yao, T., Pan, Y., Chao, H. \& Mei, T. Jointly localizing and describing events for dense video captioning. {\em Proceedings Of The IEEE Conference On Computer Vision And Pattern Recognition}. pp. 7492-7500 (2018)

\bibitem{ref17}Wang, T., Zhang, R., Lu, Z., Zheng, F., Cheng, R. \& Luo, P. End-to-end dense video captioning with parallel decoding. {\em Proceedings Of The IEEE/CVF International Conference On Computer Vision}. pp. 6847-6857 (2021)


\bibitem{ref69}Lewandowski, M., Simonnet, D., Makris, D., Velastin, S. \& Orwell, J. Tracklet reidentification in crowded scenes using bag of spatio-temporal histograms of oriented gradients. {\em Pattern Recognition: 5th Mexican Conference, MCPR 2013, Querétaro, Mexico, June 26-29, 2013. Proceedings 5}. pp. 94-103 (2013)

\bibitem{ref85}Cong, Y., Yuan, J. \& Liu, J. Abnormal event detection in crowded scenes using sparse representation. {\em Pattern Recognition}. \textbf{46}, 1851-1864 (2013)

\bibitem{ref70}Xu, J., Denman, S., Sridharan, S., Fookes, C. \& Rana, R. Dynamic texture reconstruction from sparse codes for unusual event detection in crowded scenes. {\em Proceedings Of The 2011 Joint ACM Workshop On Modeling And Representing Events}. pp. 25-30 (2011)

\bibitem{ref71}Sabokrou, M., Fathy, M., Hoseini, M. \& Klette, R. Real-time anomaly detection and localization in crowded scenes. {\em Proceedings Of The IEEE Conference On Computer Vision And Pattern Recognition Workshops}. pp. 56-62 (2015)


\bibitem{ref72}Sabokrou, M., Fayyaz, M., Fathy, M. \& Klette, R. Deep-cascade: Cascading 3d deep neural networks for fast anomaly detection and localization in crowded scenes. {\em IEEE Transactions On Image Processing}. \textbf{26}, 1992-2004 (2017)

\bibitem{ref73}Sabokrou, M., Fayyaz, M., Fathy, M., Moayed, Z. \& Klette, R. Deep-anomaly: Fully convolutional neural network for fast anomaly detection in crowded scenes. {\em Computer Vision And Image Understanding}. \textbf{172} pp. 88-97 (2018)

\bibitem{ref74}Sun, J., Wang, X., Xiong, N. \& Shao, J. Learning sparse representation with variational auto-encoder for anomaly detection. {\em IEEE Access}. \textbf{6} pp. 33353-33361 (2018)

\bibitem{ref75}Chu, W., Xue, H., Yao, C. \& Cai, D. Sparse coding guided spatiotemporal feature learning for abnormal event detection in large videos. {\em IEEE Transactions On Multimedia}. \textbf{21}, 246-255 (2018)

\bibitem{ref76}Tudor Ionescu, R., Smeureanu, S., Alexe, B. \& Popescu, M. Unmasking the abnormal events in video. {\em Proceedings Of The IEEE International Conference On Computer Vision}. pp. 2895-2903 (2017)

\bibitem{ref77}Feng, J., Zhang, C. \& Hao, P. Online learning with self-organizing maps for anomaly detection in crowd scenes. {\em 2010 20th International Conference On Pattern Recognition}. pp. 3599-3602 (2010)

\bibitem{ref78}Fan, Y., Wen, G., Li, D., Qiu, S., Levine, M. \& Xiao, F. Video anomaly detection and localization via gaussian mixture fully convolutional variational autoencoder. {\em Computer Vision And Image Understanding}. \textbf{195} pp. 102920 (2020)

\bibitem{ref79}Chalapathy, R., Menon, A. \& Chawla, S. Anomaly detection using one-class neural networks. {\em ArXiv Preprint ArXiv:1802.06360}. (2018)

\bibitem{ref80}Hasan, M., Choi, J., Neumann, J., Roy-Chowdhury, A. \& Davis, L. Learning temporal regularity in video sequences. {\em Proceedings Of The IEEE Conference On Computer Vision And Pattern Recognition}. pp. 733-742 (2016)

\bibitem{ref81}Sabokrou, M., Fathy, M. \& Hoseini, M. Video anomaly detection and localisation based on the sparsity and reconstruction error of auto-encoder. {\em Electronics Letters}. \textbf{52}, 1122-1124 (2016)


\bibitem{ref82}Wang, T., Qiao, M., Lin, Z., Li, C., Snoussi, H., Liu, Z. \& Choi, C. Generative neural networks for anomaly detection in crowded scenes. {\em IEEE Transactions On Information Forensics And Security}. \textbf{14}, 1390-1399 (2018)

\bibitem{ref67}Chandola, V., Banerjee, A. \& Kumar, V. Anomaly detection: A survey. {\em ACM Computing Surveys (CSUR)}. \textbf{41}, 1-58 (2009)

\bibitem{ref68}Chong, Y. \& Tay, Y. Modeling representation of videos for anomaly detection using deep learning: A review. {\em ArXiv Preprint ArXiv:1505.00523}. (2015)

\bibitem{ref25}Tran, D., Bourdev, L., Fergus, R., Torresani, L. \& Paluri, M. Learning spatiotemporal features with 3d convolutional networks. {\em Proceedings Of The IEEE International Conference On Computer Vision}. pp. 4489-4497 (2015)

\bibitem{ref35}Carreira, J. \& Zisserman, A. Quo vadis, action recognition? a new model and the kinetics dataset. {\em Proceedings Of The IEEE Conference On Computer Vision And Pattern Recognition}. pp. 6299-6308 (2017)

\bibitem{ref36}Radford, A., Kim, J., Hallacy, C., Ramesh, A., Goh, G., Agarwal, S., Sastry, G., Askell, A., Mishkin, P., Clark, J. \& Others Learning transferable visual models from natural language supervision. {\em International Conference On Machine Learning}. pp. 8748-8763 (2021)

\bibitem{ref43}Vaswani, A., Shazeer, N., Parmar, N., Uszkoreit, J., Jones, L., Gomez, A., Kaiser, Ł. \& Polosukhin, I. Attention is all you need. {\em Advances In Neural Information Processing Systems}. \textbf{30} (2017)

\bibitem{ref42}Zhu, X., Su, W., Lu, L., Li, B., Wang, X. \& Dai, J. Deformable detr: Deformable transformers for end-to-end object detection. {\em ArXiv Preprint ArXiv:2010.04159}. (2020)

\bibitem{ref83}Wang, C., Bochkovskiy, A. \& Liao, H. YOLOv7: Trainable bag-of-freebies sets new state-of-the-art for real-time object detectors. {\em Proceedings Of The IEEE/CVF Conference On Computer Vision And Pattern Recognition}. pp. 7464-7475 (2023)

\bibitem{ref84}Du, Y., Zhao, Z., Song, Y., Zhao, Y., Su, F., Gong, T. \& Meng, H. Strongsort: Make deepsort great again. {\em IEEE Transactions On Multimedia}. (2023)

\bibitem{ref55}Sun, X., Gao, J., Zhu, Y., Wang, X. \& Zhou, X. Video Moment Retrieval via Comprehensive Relation-aware Network. {\em IEEE Transactions On Circuits And Systems For Video Technology}. pp. 1-1 (2023)

\bibitem{ref61}Zhang, G., Zhang, H., Lin, W., Chandran, A. \& Jing, X. Camera Contrast Learning for Unsupervised Person Re-Identification. {\em IEEE Transactions On Circuits And Systems For Video Technology}. pp. 1-1 (2023)

\bibitem{ref56}Zhang, J., Xie, Y., Ding, W. \& Wang, Z. Cross on Cross Attention: Deep Fusion Transformer for Image Captioning. {\em IEEE Transactions On Circuits And Systems For Video Technology}. pp. 1-1 (2023)

\bibitem{ref58}Yang, X., Lv, F., Liu, F. \& Lin, G. Self-Training Vision Language BERTs with a Unified Conditional Model. {\em IEEE Transactions On Circuits And Systems For Video Technology}. pp. 1-1 (2023)


\bibitem{ref47}Carion, N., Massa, F., Synnaeve, G., Usunier, N., Kirillov, A. \& Zagoruyko, S. End-to-end object detection with transformers. {\em Computer Vision–ECCV 2020: 16th European Conference, Glasgow, UK, August 23–28, 2020, Proceedings, Part I 16}. pp. 213-229 (2020)

\bibitem{ref49}Rezatofighi, H., Tsoi, N., Gwak, J., Sadeghian, A., Reid, I. \& Savarese, S. Generalized intersection over union: A metric and a loss for bounding box regression. {\em Proceedings Of The IEEE/CVF Conference On Computer Vision And Pattern Recognition}. pp. 658-666 (2019)

\bibitem{ref48}Lin, T., Goyal, P., Girshick, R., He, K. \& Dollár, P. Focal loss for dense object detection. {\em Proceedings Of The IEEE International Conference On Computer Vision}. pp. 2980-2988 (2017)

\bibitem{ref51}Zhu, X., Su, W., Lu, L., Li, B., Wang, X. \& Dai, J. Deformable detr: Deformable transformers for end-to-end object detection. {\em ArXiv Preprint ArXiv:2010.04159}. (2020)

\bibitem{ref20}Papineni, K., Roukos, S., Ward, T. \& Zhu, W. Bleu: a method for automatic evaluation of machine translation. {\em Proceedings Of The 40th Annual Meeting Of The Association For Computational Linguistics}. pp. 311-318 (2002)

\bibitem{ref21}Banerjee, S. \& Lavie, A. METEOR: An automatic metric for MT evaluation with improved correlation with human judgments. {\em Proceedings Of The Acl Workshop On Intrinsic And Extrinsic Evaluation Measures For Machine Translation And/or Summarization}. pp. 65-72 (2005)

\bibitem{ref23}Vedantam, R., Lawrence Zitnick, C. \& Parikh, D. Cider: Consensus-based image description evaluation. {\em Proceedings Of The IEEE Conference On Computer Vision And Pattern Recognition}. pp. 4566-4575 (2015)

\bibitem{ref22}Lin, C. Rouge: A package for automatic evaluation of summaries. {\em Text Summarization Branches Out}. pp. 74-81 (2004)

\bibitem{ref24}Fujita, S., Hirao, T., Kamigaito, H., Okumura, M. \& Nagata, M. SODA: Story oriented dense video captioning evaluation framework. {\em Computer Vision–ECCV 2020: 16th European Conference, Glasgow, UK, August 23–28, 2020, Proceedings, Part VI 16}. pp. 517-531 (2020)

\bibitem{ref26}Karpathy, A., Toderici, G., Shetty, S., Leung, T., Sukthankar, R. \& Fei-Fei, L. Large-scale video classification with convolutional neural networks. {\em Proceedings Of The IEEE Conference On Computer Vision And Pattern Recognition}. pp. 1725-1732 (2014)

\bibitem{ref27}Kingma, D. \& Ba, J. Adam: A method for stochastic optimization. {\em ArXiv Preprint ArXiv:1412.6980}. (2014)

\bibitem{ref57}Li, H., Liu, M., Hu, Z., Nie, F. \& Yu, Z. Intermediary-guided Bidirectional Spatial-Temporal Aggregation Network for Video-based Visible-Infrared Person Re-Identification. {\em IEEE Transactions On Circuits And Systems For Video Technology}. pp. 1-1 (2023)

\bibitem{yang2023vid2seq}Yang, A., Nagrani, A., Seo, P., Miech, A., Pont-Tuset, J., Laptev, I., Sivic, J. \& Schmid, C. Vid2seq: Large-scale pretraining of a visual language model for dense video captioning. {\em Proceedings Of The IEEE/CVF Conference On Computer Vision And Pattern Recognition}. pp. 10714-10726 (2023)

\bibitem{ref30}Xu, H., Ye, Q., Yan, M., Shi, Y., Ye, J., Xu, Y., Li, C., Bi, B., Qian, Q., Wang, W. \& Others mplug-2: A modularized multi-modal foundation model across text, image and video. {\em ArXiv Preprint ArXiv:2302.00402}. (2023)


\bibitem{ref31}Chen, S., Li, H., Wang, Q., Zhao, Z., Sun, M., Zhu, X. \& Liu, J. VAST: A Vision-Audio-Subtitle-Text Omni-Modality Foundation Model and Dataset. {\em ArXiv Preprint ArXiv:2305.18500}. (2023)


\bibitem{ref32}Wang, J., Yang, Z., Hu, X., Li, L., Lin, K., Gan, Z., Liu, Z., Liu, C. \& Wang, L. Git: A generative image-to-text transformer for vision and language. {\em ArXiv Preprint ArXiv:2205.14100}. (2022)


\bibitem{ref33}He, X., Chen, S., Ma, F., Huang, Z., Jin, X., Liu, Z., Fu, D., Yang, Y., Liu, J. \& Feng, J. VLAB: Enhancing Video Language Pre-training by Feature Adapting and Blending. {\em ArXiv Preprint ArXiv:2305.13167}. (2023)



\bibitem{zhou2018end}Zhou, L., Zhou, Y., Corso, J., Socher, R. \& Xiong, C. End-to-end dense video captioning with masked transformer. {\em Proceedings Of The IEEE Conference On Computer Vision And Pattern Recognition}. pp. 8739-8748 (2018)

\bibitem{iashin2020better}Iashin, V. \& Rahtu, E. A better use of audio-visual cues: Dense video captioning with bi-modal transformer. {\em ArXiv Preprint ArXiv:2005.08271}. (2020)



%

\end{thebibliography}
\end{document}